\documentclass{ecai} 

\usepackage{multirow}
\usepackage{latexsym}
\usepackage{amssymb}
\usepackage{amsmath}
\usepackage{amsthm}
\usepackage{booktabs}
\usepackage{enumitem}
\usepackage{graphicx}
\usepackage{hyperref}
\usepackage{color}
\usepackage{algorithm}
\usepackage[noend]{algpseudocode}

\usepackage[table,xcdraw]{xcolor}

\newcommand{\BibTeX}{B\kern-.05em{\sc i\kern-.025em b}\kern-.08em\TeX}

\begin{document}


\begin{frontmatter}


\paperid{2337} 


\title{Multi-agent Planning using Visual Language
Models}


\author[*]{\fnms{Michele}~\snm{Brienza}\orcid{0009-0000-1549-9500}\footnote{Corresponding Author. Email: \{brienza, argenziano\}@diag.uniroma1.it, vincenzo.suriani@unibas.it, domenico.bloisi@unint.eu}}
\author[*]{\fnms{Francesco}~\snm{Argenziano}\orcid{0009-0004-2028-7253}\footnotemark}
\author[†]{\fnms{Vincenzo}~\snm{Suriani}\orcid{0000-0003-1199-8358}\footnotemark} 
\author[§]{\fnms{Domenico D.}~\snm{Bloisi}\orcid{0000-0003-0339-8651}\footnotemark} 
\author[*]{\fnms{Daniele}~\snm{Nardi}\orcid{0000-0001-6606-200X}}

\address[*]{Sapienza University of Rome, Rome RM 00181, Italy}
\address[†]{University of Basilicata, Potenza PZ 85100, Italy}
\address[§]{International University of Rome UNINT, Rome RM 00147, Italy}


\begin{abstract}
Large Language Models (LLMs) and Visual Language Models (VLMs) are attracting increasing interest due to their improving performance and applications across various domains and tasks. However, LLMs and VLMs can produce erroneous results, especially when a deep understanding of the problem domain is required. For instance, when planning and perception are needed simultaneously, these models often struggle because of difficulties in merging multi-modal information. To address this issue, fine-tuned models are typically employed and trained on specialized data structures representing the environment. This approach has limited effectiveness, as it can overly complicate the context for processing. In this paper, we propose a multi-agent architecture for embodied task planning that operates without the need for specific data structures as input. Instead, it uses a single image of the environment, handling free-form domains by leveraging commonsense knowledge. We also introduce a novel, fully automatic evaluation procedure, PG2S, designed to better assess the quality of a plan. We validated our approach using the widely recognized ALFRED dataset, comparing PG2S to the existing KAS metric to further evaluate the quality of the generated plans.
\end{abstract}

\end{frontmatter}


\section{Introduction}

Foundation Models (FMs) are machine learning models that are trained on a broad (\textit{Internet-scale}) amount of data and can be refined to be used in a wide range of downstream applications \citep{bommasani2021opportunities}. Initial examples of these models, i.e., Large Language Models (LLMs) \citep{devlin2018bert,brown2020language,achiam2023gpt,touvron2023llama}, 
were inherently of the Natural Language Processing (NLP) field. Nevertheless, in the last years, we have witnessed the emergence of multi-modal LLMs, which can handle non-textual inputs and outputs.
Visual Language Models (VLMs) \citep{liu2023visual,ramesh2021zeroshot} have particular relevance in this category since they can take as input images and/or textual queries and generate contextual high-quality outputs. Additionally, the birth of many toolkits like HuggingFace \citep{wolf2019huggingface} or LangChain \citep{10242497} have contributed to the outburst and the distribution of such models, 
widening their domain of applications.

It has been demonstrated that LLMs can be used as zero-shot \citep{pmlr-v162-huang22a} and few-shot \citep{song2023llmplanner} planners.
This is due to the fact that these models have been trained on huge amounts of data, therefore they incorporate the commonsense knowledge proper of humans \citep{li-etal-2022-systematic}.

An agent with commonsense knowledge acquires complex reasoning capabilities via chain-of-thought \citep{wei2023chainofthought} and it becomes able to correctly generate a plan to achieve the desired goal.
The generated plans are grounded in the sense that actions, objects, and states all refer to the specific environment the embodied agent is deliberating in, thanks to the information incorporated in the queries.
\begin{figure}[t!]
\centering
\includegraphics[width=0.8\columnwidth]{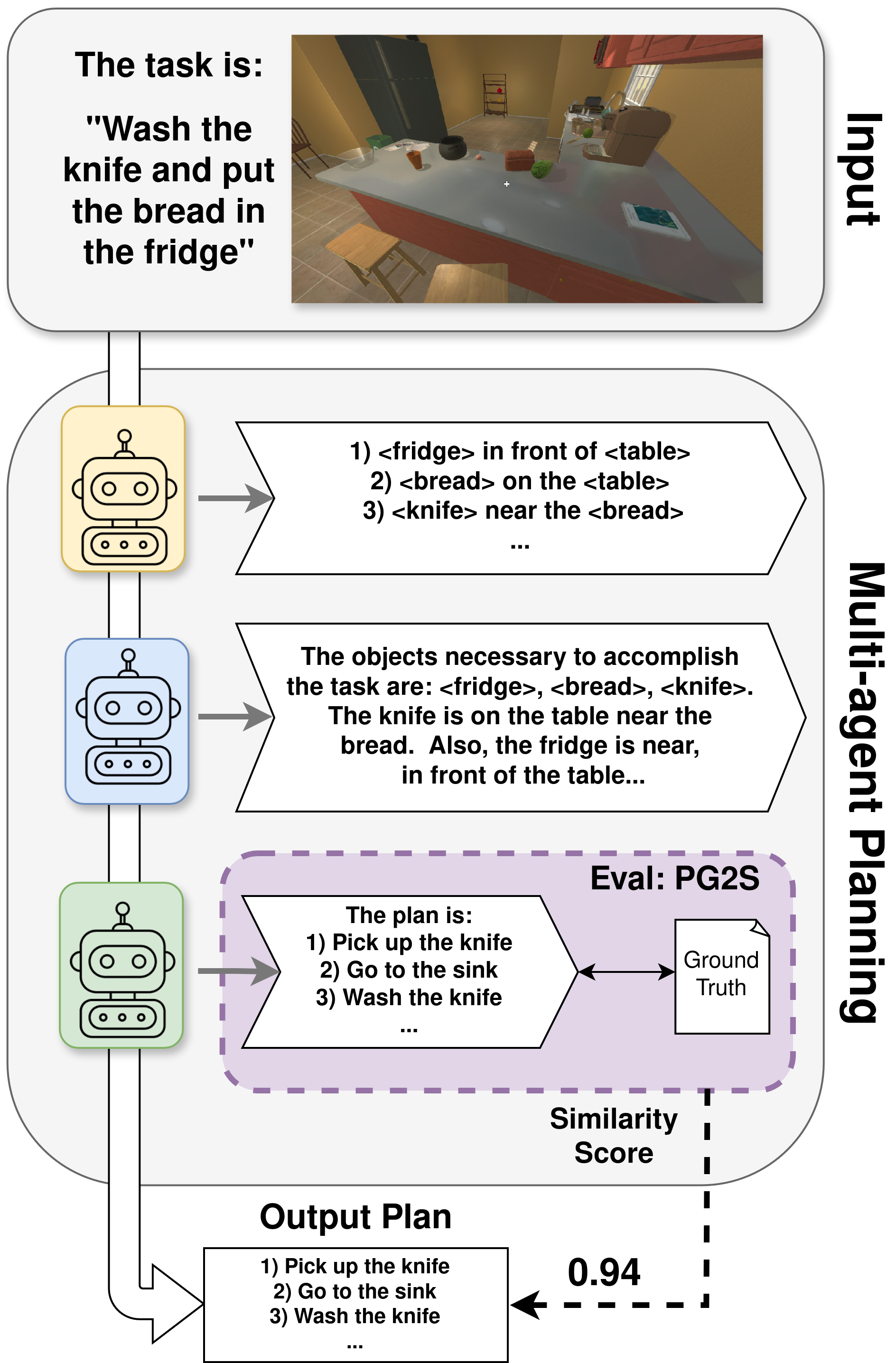}

\caption{Overall view of the proposed framework. Given a task description and an image of the scene, the plan is obtained with multi-agent planning and assessed with the new score.}

\label{fig:1}
\end{figure}
\begin{figure*}[t!]
\centering
\includegraphics[width=17.5cm]{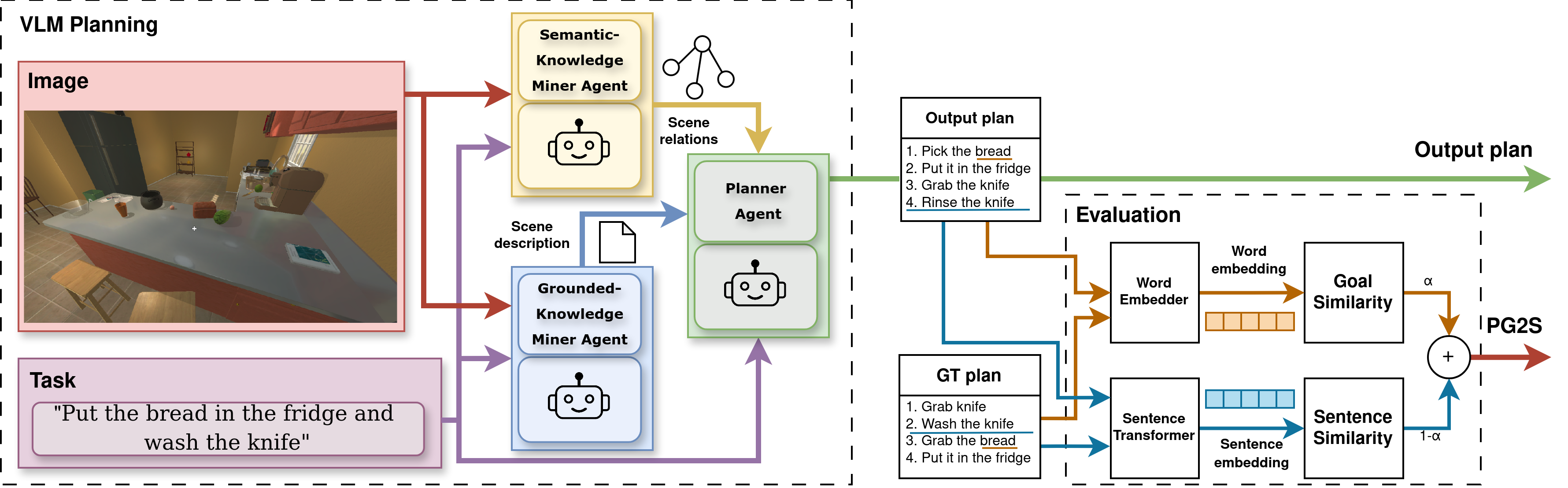}
\caption{Complete and detailed architecture of the proposed method. The task description and the image are given in input to the agents that extract meaningful information from the scene. Their output is then processed by the planner agent that obtain the final plan. Such plan is then compared with the ground truth  and evaluated according our new metric that takes into account semantically meaningful information.}
\label{fig:example2}
\end{figure*}
Existing solutions for grounding concern encoding the environment in a structured manner, i.e., using tables or graphs \citep{lin2023grounded,rana2023sayplan}, since they are easily promptable to the model once converted in some sort of streamable format.
However, these kinds of representations grow very quickly as the environment grows in size, thus it becomes difficult to incorporate them in a language model query prompt. The context window, namely the amount of text the model can handle as input when generating language, becomes very big as the prompt increases, and the output could be affected by several hallucinations \cite{liu2024lost}. This is a problem in any LLMs application, but particularly when we are trying to plan a specific procedure to achieve a certain goal.
For this reason, it is important to keep the input to the model as small as possible, 
including only the necessary information to carry out the desired task. In fact, decomposing the goal into several sub-goals for multiple independent agents can drastically improve the final output, thus performing better w.r.t. single-agent architectures \citep{wang2023unleashing,talebirad2023multi,musumeci2024llm}.

In this paper, we analyze the capabilities of FMs when used as reasoning components. In particular, we use FMs to deploy grounded plans for embodied agents in free-form domains, i.e. domains that come without a structured representation. All we need is a picture of the scene that captures the most relevant aspects of the environment, and a textual query on the goal we want to achieve.  Our approach exploits a hierarchical multi-agent structure, in the sense that every agent is a different VLM/LLM instance which addresses to solve only one aspect of the whole planning procedure, according to the definition given by \cite{talebirad2023multi}. In this way, every agent has a limited context window, thus being less prone to hallucinations.

We conduct a comparative study to demonstrate that our single-image multi-agent scheme outperforms both an architecture utilizing a structured environment representation (such as tabular) and a single-agent architecture where all input is fed to the VLM simultaneously.

We also propose \emph{Planning Goal Semantic Score} (PG2S), a new metric that does not rely on user validations to evaluate the results. PG2S does not consider the partial ordering of actions needed to carry out the plan to achieve a goal, and it is semantically sound in the sense that it deals with synonyms without losing the meaning of the plan.
In Fig. \ref{fig:1} it is possible to see an overall view of our system. 
In summary, the contribution of this work is three-fold:
\begin{itemize}
    \item We demonstrate that it is possible to soundly plan to achieve a task with a VLM and a query input, relaxing the assumptions about complex structured query;
    \item We propose a multi-agent framework to decompose the final task into different sub-tasks, thus reducing the risk of hallucinations and other harmful phenomena;
    \item We introduce a new metric, \emph{PG2S}, to autonomously evaluate the correctness of a plan expressed in natural language that is partial-ordering agnostic and semantically aware.
\end{itemize}
We validate our approach on ALFRED  \citep{shridhar2020alfred}, a benchmark purposefully designed to evaluate natural language instructions mapping in household
environments, built upon the AI2-THOR framework \citep{kolve2022ai2thor}. 
For comparison, we use G-PlanET \citep{lin2023grounded}, which contains all the simulation environments of ALFRED but represented in a tabular form. \\
We release the code, the prompts used and the results obtained on our project website.\footnote{\url{https://lab-rococo-sapienza.github.io/map-vlm/}}


\section{Related Work}
In this section, we discuss existing solutions about both using LLMs for planning and adopting a multi-agent architecture for prompting.  

\subsection{LLMs as Planners}
A pioneering work that exploits the use of LLMs for embodied agents is SayCan \citep{ahn2022can}, where a robot can behave as ``hands and eyes'' for an LLM when grounding tasks in real-world scenarios, taking advantage of the semantic knowledge of the model when performing complex instructions. Following this research, several approaches started to emerge that tried to use LLMs as the planning component 
in many different use cases.

Huang et al. \citep{pmlr-v162-huang22a} demonstrate that LLMs behave like zero-shot planners when they are correctly prompted. In contrast, Song et al. \citep{song2023llmplanner} show that tuning these models in a few-shot setting, allows them to surpass state-of-the-art Vision Language Navigation (VLN) models even if they are trained on a broader amount of data, thanks to LLMs' embedded commonsense knowledge.

LLMs' capabilities change when the query in input is not completely textual, but can assume a more structured form, e.g., a tabular structure \citep{lin2023grounded}, a graph-like structure \citep{rana2023sayplan} (such as 3D Scene Graphs \citep{armeni20193d}), or even LTL formulas \citep{dai2023optimal}. Incorporating this additional information is useful to improve the overall performance in the desired tasks. However, the biggest drawback of these techniques is that they require a very high computational cost when applied to real-world scenarios, where the environment is unstructured.

New research directions were possible thanks to advancements in VLMs, such as the possibility of directly processing visual queries given to embodied agents. Allowing systems to take in input images of the environment can solve the problem of creating complex structured representations, saving time and computations while still maintaining the reasoning power of LLMs. 
For example, \citep{dorbala2023can} shows that it was possible to use VLMs to find objects in the environment that are described by natural language descriptions given by humans (like a ``cat-shaped mug''). 

However, choosing which VLM to use is not a trivial task. Many VLMs are built upon CLIP \citep{radford2021learning}, but these models present \textit{bags-of-words} behaviors \citep{yuksekgonul2022and}, namely they ignore semantic structures of inputs, such as spatial relationships between objects. To cope with this problem, LLM-Grounder \citep{yang2023llm} shows that it was indeed possible to exploit the power of VLMs to plan for embodied agents while reducing the effects of the \textit{bags-of-words} phenomenon. In our approach, we adopt VLMs to get rid of complex structured inputs, but at the same time, we reduce as much as possible the \textit{bags-of-words} behavior by using a multi-agent approach. Our approach is able to decompose the task into sub-tasks in such a way that potentially problematic relations are handled by a specific agent.

\subsection{Multi-agent Prompting}
As LLMs became more and more diffused, it was discovered that specific prompting patterns produced better results than free-form prompts (\textit{prompt engineering}) \citep{zamfirescu2023johnny}. In planning applications, \textit{chain-of-thought} reasoning \citep{wei2023chainofthought} has marked a notable advance, with multi-step reasoning.

Another important step in prompt engineering with LLMs is achieved by leveraging the power of multi-agent systems. In \citep{talebirad2023multi}, a collaborative environment where multiple agents with different roles had to work together to accomplish a task, is demonstrated to have better performance w.r.t. a single-agent. Moreover, results improved not only in settings with many role-specific agents but also in settings with multi-persona self-collaborating agents\citep{wang2023unleashing}. 

Several frameworks started to emerge, simplifying the development of multi-agent applications \citep{wu2023autogen,rasal2024llm,shanahan2023role}. 
As a drawback, these frameworks intrinsically increase the complexity of the systems that adopt them.


\section{Methodology}
\noindent
The typical interaction between an LLM and a user consists of a trial-and-error process to obtain the desired result by refining the prompt. The accuracy of the environmental information is crucial to obtain a correct plan. 
Usually, this information comes from tables or structured data. 
Our method is based on relaxing the structured information known a priori from a previous labeling process. 
In our architecture, we use a multi-agent pipeline that takes as input only an image of the environment, along with the task to execute. Then, we show how this strategy allows us to have a correct plan, even in free-form domains.
To assess the correctness, we use our PG2S metric by comparing the plans devised from images and those from tables by referring to ALFRED's annotations. 

\subsection{Multi-agent Planning} 
Our solution employs three agents, each representing a phase in the planning generation process: the \emph{Semantic-Knowledge Miner Agent} (SKM), the \emph{Grounded-Knowledge Miner Agent} (GKM), and the \emph{Planner Agent} (\textit{P}). GPT-4V is used for agents that process images, while GPT-4 is used for the planning agent \citep{achiam2023gpt}.

The SKM Agent identifies object classes within the image and establishes the scene's ontology. It also determines relationships between objects, creating a knowledge graph. The GKM Agent grounds these objects, providing short descriptions that include their relationships with surrounding objects, resulting in a high-level yet structurally sound scene description. The \textit{P} Agent then generates a plan using the information from the SKM and GKM Agents. This method minimizes hallucinations and focuses the plan on the relevant objects in the scene.

Using a Visual Language Model (VLM), we achieve better results with a multi-agent strategy compared to a single-agent approach. In a single-agent setup, the prompt directs the VLM to create a plan from the input image. In contrast, the multi-agent setup allows the Miner Agents to enrich the Planning Agent’s knowledge with detailed environmental information, as illustrated in Fig. \ref{fig:example2}.

The multi-agent strategy enhances plan quality by distributing the workload among agents, each handling specific tasks. This division reduces the risk of hallucinations by maintaining smaller, more focused prompts within each agent’s context window \cite{liu2024lost}. By splitting the task into simpler sub-tasks, our pipeline ensures more accurate and coherent responses, following the "divide and conquer" principle.

\subsection{Evaluation}
Choosing an adequate metric to evaluate the quality of produced plans is not trivial. Usually, only the Success Rate (SR) or the SR weighted by the inverse path length (SRL) are used to evaluate the plan correctness \citep{song2023llmplanner,dorbala2023can}. However, these metrics are not very convenient to compute, and researchers often rely on Amazon Mechanical Turk to check the correctness using human experts. Moreover, they do not evaluate the quality of the plan: they state how many times the goal is achieved and how the length of the plan influences the result.

G-PlanET \citep{lin2023grounded} tries to define a new metric to cope with this problem: inspired by metrics used for semantic captioning like CIDEr \citep{vedantam2015cider} and SPLICE \citep{anderson2016spice}, it proposes KeyActionScore (KAS). KAS builds a set of key action phrases obtained from every step of the generated plan $\hat{S}_i$, and from the reference plan of the dataset $S_i$. Then, by checking how many action phrases in $\hat{S}_i$ are covered by $S_i$, and by computing this precision, it is possible to evaluate the matching quality of the two sets for the $i$-th step of the plan.

This metric present two main limitations.
The first is that it always assumes that the reference plan is correct, which is not always true as we found some examples in the ALFRED dataset of plans that are not completely correct: e.g. the reference plan for the goal \textit{"Put a hot bread in the refrigerator"} has as one of the steps the action \textit{"put the knife in the microwave"} which is extremely dangerous and globally incorrect for the desired goal.
The second is that in definition of KAS, a mapping is considered correct \textit{if and only if} it follows the order of actions given by the step. This is a strong assumption, since there are many plans in which the order of actions is not necessary to reach a goal \citep{minton1994total}, so it can penalize plans that are actually correct. 

\noindent
To this end, we propose a new metric, \textbf{PG2S}, that copes with this problem.
As an example, we show a reference plan that can be used as a ground truth plan and a possible predicted plan (see Table \ref{esempio}).
\begin{table}[t]
\caption{Example of partial-ordering of actions.}
\begin{tabular}{ll}
\begin{tabular}[c]{@{}l@{}}\textbf{Ground truth plan: }\\Carefully wear the left sock.\\ Wear the right sock.\\ Put on the snug left shoe carefully.\\ Slip into the right shoe comfortably.\end{tabular} & \begin{tabular}[c]{@{}l@{}}\textbf{Predicted plan: }\\Wear the right sock with care.\\ Carefully wear the right shoe.\\ Gently wear the left sock first.\\ Slide into the left shoe carefully.\end{tabular}    
\end{tabular}
\label{esempio}
\end{table}
The predicted plan to reach the goal \textit{"Wear a pair of shoes"} is correct for a human evaluator. Despite this, the plan is different from the ground truth in the order of the actions, and the evaluation should be able to take into account this possibility. Using the KAS metric the similarity score is equal to $0.33$; while for PG2S (ours) the similarity score obtained is equal to $0.83$. Algorithm \ref{alg1} presents the procedure used to compute such an evaluation score.
More in detail, given two sets of planning descriptions, \( \mathcal{P}_{gt} \) and \( \mathcal{P}_{pred} \), respectively the ground truth plan and the predicted plan, we aim at quantifying their similarity, using two levels of evaluation, namely a \textit{sentence-wise} and a \textit{goal-wise}, both based on the semantic values. To determine if two embeddings are similar we use a threshold mechanism. In particular, we adopt the approach presented in \citep{rekabsaz2017exploration}, where the authors obtain thresholds that vary according to the dimensionality of the embedding vector and verify that their use allows to obtain only semantically similar elements.
\begin{algorithm}[t!]
    \caption{PG2S Evaluation Procedure}\label{alg1}
    \begin{algorithmic}[1]
        \Require $\mathcal{P}_{gt} \,\text{ground truth plan}, \mathcal{P}_{pred} \,\text{predicted 
        plan}$
        \Ensure $PG2S$
        
        \State $MaxSimilPlan, MaxSimilGoal \gets [\:]$
        \For{$s_i \in \mathcal{P}_{gt}$}
            \State find the most similar sentence $s_j$ in $\mathcal{P}_{pred}$ \label{lst:line:line3}
            \State \textbf{if} exists: add $1$ to $MaxSimilPlan$; \textbf{otherwise} add $0$
            \State $\mathcal{P}_{pred}$.pop($s_j$)\label{lst:line:line5}
        \EndFor
        \State $S_{plan} \gets \text{mean}(MaxSimilPlan)$\label{lst:line:line6}
        \State $A_{gt}, A_{pred} \gets [\:]$
        \For{$s_i , s_j \in \mathcal{P}_{gt},  \mathcal{P}_{pred} $}
            \State add actions in $A_{gt}$ and $A_{pred}$ with $Framing()$\label{lst:line:line7}
        \EndFor
        \For{$a_i \in A_{gt}$}
                \State find the most similar action $a_j$ in $\mathcal{A}_{pred}$ \label{lst:line:line8}
                \State \textbf{if} exists: add $1$ to $MaxSimilGoal$; \textbf{otherwise} add $0$
            \State $\mathcal{A}_{pred}$.pop($a_j$)\label{lst:line:line13}
        \EndFor
        \State $S_{goal} \gets \text{mean}(MaxSimilGoal)$\label{lst:line:line14}
        \State $PG2S \gets \alpha * S_{plan} + (1-\alpha)*S_{goal}$
    \end{algorithmic}
    
\end{algorithm}

\textit{\textbf{Sentence-wise similarity.}}
To compute the sentence similarity, we deploy embedding vector representations
for each sentence using a \emph{Sentence Transformer}. In particular, we use MPNet \citep{song2020mpnet}, which achieves better results in semantic evaluation tasks compared with previous state-of-the-art pre-trained models \citep{song2020mpnet} (e.g., BERT, XLNet, and RoBERTa).
For each sentence \( s_i \in \mathcal{P}_{gt} \) and \( s_j \in \mathcal{P}_{pred} \), we obtain the similarity between their embeddings (${v}_i$ and ${v}_j$) using the cosine similarity $cos(v_i,v_j)$.
For each \( s_i \), we identify the most similar sentence in \( \mathcal{P}_{pred} \) (line~\ref{lst:line:line3}) and remove it from the set (line~\ref{lst:line:line5}). The value of each similarity yields a list of maximum similarity scores. The sentence-wise similarity is the average of these scores (line~\ref{lst:line:line6}).
\begin{equation}
S_{\text{plan}} (\mathcal{P}_{gt}, \mathcal{P}_{pred}) = \frac{1}{N}\sum_{i=1}^{N}  \text{MaxSimilPlan}_i
\end{equation}
\noindent 

\textit{\textbf{Goal-wise similarity.}}
To compute the goal similarity, we first perform a POS tagging pre-processing stage using \textit{spaCy} \citep{spacy2}, and then, for each sentence we extract the main action using a \emph{Framing()} procedure (line~\ref{lst:line:line7}). This procedure works as follows:
for each word in a sentence, we add it in the action set if it is either \textit{i.} a central (`root') verb (VERB), or \textit{ii.} if it is a noun (NOUN) and its dependency tag is either a `direct object' (DOBJ) or the `nominal subject' (NSUBJ). In this way, for each step we obtain the main action and the involved objects.
For each action $a_i \in \mathcal{A}_{gt}$ and $a_j \in \mathcal{A}_{pred}$, we obtain a similarity value from the product between the mean of nouns similarity and the verbs similarity, obtained from a \textit{WordEmbeddingSimilarity()} tool (Word2Vec \citep{mikolov2013efficient}).We consider two nouns and two verbs to be similar if their similarity value exceeds a threshold $\tau=0.708$, according to \cite{rekabsaz2017exploration}.

For each action \( a_i \) in \( \mathcal{A}_{gt} \) we identify the most similar action in \( \mathcal{A}_{pred} \) and remove it from the set. The most similar action is found using the combined similarity computed with the product of both values (line~\ref{lst:line:line8}) and removed from \( \mathcal{A}_{pred} \) (line~\ref{lst:line:line13}).
The value of each action similarity yields a list of maximum similarity scores. The average of these scores gives us the goal-wise similarity of the sets (line~\ref{lst:line:line14}).
\begin{equation}
S_{\text{goal}} (\mathcal{A}_{gt}, \mathcal{A}_{pred}) = \frac{1}{N}\sum_{i=1}^{N}  \text{MaxSimilGoal}_i
\end{equation}
\noindent 

\textit{{\textbf{PG2S.}}}
The final similarity score is our metric PG2S, which is a weighted average of the sentence-level and action-state similarities, where \( \alpha \) is a weighting factor, set to $0.5$ to equally balance the contributions of the two scores:
\begin{equation}
PG2S = (1-\alpha)*S_{\text{plan}} (\mathcal{P}_{gt}, \mathcal{P}_{pred}) + \alpha*S_{\text{goal}}(\mathcal{A}_{gt}, \mathcal{A}_{pred})
\end{equation}
Another issue arises because KAS employs a set intersection, whereby terms that are not equal are not considered for the similarity calculation. This can result in the problem of having the same action with a subject that is not appropriate for use in the case of goal similarity. To illustrate this aspect, consider the action "Walk to the desk" in comparison to "Walk to the moon". In the case of KAS, the resulting similarity score is 0.67 because two out of three elements are equal, whereas in PG2S, the similarity score is 0. This discrepancy can be attributed to the fact that KAS does not consider the nuances of natural language, whereas PG2S does. 



\section{Experimental Results}

This section presents the outcomes of the conducted experiments, which were designed to test the proposed architecture's validity. The results obtained using a single image are presented and then compared with a structured perception of the environment, as seen in state-of-the-art works. The output plans regarding home scenarios tasks are taken from the ALFRED dataset using the AI2Thor environment.
Chosen the image and the environment, for each of those we have found the plan associated with the scene and saved the ground truth plans that we have used to compare our results. The environment scenarios are chosen by selecting several different situations in order to have various complexity and domains of application according to the chosen fields of ALFRED such as: picking up objects and placing them; picking up objects, heating or cooling them, and place them somewhere else; cleaning objects and examining under the light; and more.
\subsection{Evaluation of our PG2S Metric}
\vspace{-0.1cm}
During the experimental phase of PG2S development, a series of tests were conducted to ensure the correctness of the metric. 
Specifically, we compared ALFRED plans with those predicted by our architecture, together with their corrupted version.
During the test phase, several examples were selected from the ALFRED dataset. The plans obtained were checked qualitatively and it was possible to verify that the plans generated by the multi-agent architecture are correct in terms of the sequence of actions. The corrupted plans ensure that the goal similarity will be respected and will not return a high level of similarity in case of ambiguity.
\begin{table*}[h!]
\caption{Similarity values are calculated between predicted plans and ALFRED-annotated plans, where the predicted plans may be corrupted by substituting object names in the steps. For corrupted plans, lower similarity values are preferred because they indicate a greater difference from the ground truth plan. Conversely, for uncorrupted plans, higher similarity values are desirable.}
\centering
\begin{tabular}{|
>{}c |
>{}c |
>{}c |
>{}c |
>{}c |}
\hline
\textbf{Corrupted plan}                               & \multicolumn{1}{r|}{\textbf{PG2S}} & \textbf{KAS}  & \textbf{PG2S not corrupted} & \textbf{KAS not corrupted}\\ \hline
{ {trial\_T20190909\_075240\_427378} (laptop $\rightarrow$ bread; pen $\rightarrow$ knife)  } & \textbf{0.138}                                                      & 0.148  & \textbf{0.458} & 0.211     \\ \hline
{trial\_T20190906\_185208\_580877} (bathroom $\rightarrow$ kitchen; bottle $\rightarrow$ tomato) & \textbf{0.167} & 0.260 & \textbf{0.417} & 0.302 \\ \hline{trial\_T20190907\_020543\_865134} (monitor $\rightarrow$ statue; fire $\rightarrow$ lamp)   &\textbf{ 0.000 } & 0.208 & \textbf{0.500}  & 0.311\\ \hline{trial\_T20190907\_143702\_923249}  (moon $\rightarrow$ desk; mouse $\rightarrow$ card)  & \textbf{0.000}  & \textbf{0.000} & \textbf{0.875}  & 0.090\\ \hline{trial\_T20190907\_171916\_941174} (pizza $\rightarrow$ coffee)  & 0.083 & \textbf{0.055} & \textbf{0.167} & 0.104\\ \hline
{trial\_T20190909\_035341\_047789} (card $\rightarrow$ pencil; restaurant $\rightarrow$ desk)  & \textbf{0.000}   & 0.036 & \textbf{0.500}   & 0.102 \\ \hline
\end{tabular}%
\label{tab:results_corrupted}
\end{table*}

\begin{table*}[h!]

\caption{Similarity values among predicted and ALFRED annotated plans using thirty plans in ten different scenes using tables and ground truth plans by the G-PLANET dataset.}
\centering
\begin{tabular}{|c|cc|cc|cc|cc|}
\hline
\multirow{2}{*}{\textbf{Task\_ID}} &
  \multicolumn{2}{l|}{\textbf{Single-agent w/ table}} &
  \multicolumn{2}{l|}{\textbf{Multi-agent w/ table}} &
  \multicolumn{2}{l|}{\textbf{Single-agent w/ image}} &
  \multicolumn{2}{l|}{\textbf{Multi-agent w/ image}} \\ \cline{2-9} 
 &
  \multicolumn{1}{c|}{\textbf{PG2S}} &
  \textbf{KAS} &
  \multicolumn{1}{c|}{\textbf{PG2S}} &
  \textbf{KAS} &
  \multicolumn{1}{c|}{\textbf{PG2S}} &
  \textbf{KAS} &
  \multicolumn{1}{c|}{\textbf{PG2S}} &
  \textbf{KAS} \\ \hline
trial\_T20190907\_161326\_928347 &
  \multicolumn{1}{c|}{0.00} &
  \textbf{0.25} &
  \multicolumn{1}{c|}{0.00} &
  \textbf{0.31} &
  \multicolumn{1}{c|}{\textbf{0.30}} &
  0.11 &
  \multicolumn{1}{c|}{\textbf{0.10}} &
  0.05 \\ \hline
trial\_T20190910\_173916\_331859-1 &
  \multicolumn{1}{c|}{\textbf{0.14}} &
  None &
  \multicolumn{1}{c|}{\textbf{0.14}} &
  None &
  \multicolumn{1}{c|}{\textbf{0.39}} &
  0.15 &
  \multicolumn{1}{c|}{\textbf{0.43}} &
  0.19 \\ \hline
trial\_T20190909\_004531\_429065-1 &
  \multicolumn{1}{c|}{\textbf{0.14}} &
  None &
  \multicolumn{1}{c|}{\textbf{0.14}} &
  None &
  \multicolumn{1}{c|}{\textbf{0.29}} &
  None &
  \multicolumn{1}{c|}{\textbf{0.21}} &
  None \\ \hline
trial\_T20190907\_114323\_767231-1 &
  \multicolumn{1}{c|}{\textbf{0.10}} &
  None &
  \multicolumn{1}{c|}{0.30} &
  \textbf{0.32} &
  \multicolumn{1}{c|}{\textbf{0.80}} &
  0.38 &
  \multicolumn{1}{c|}{\textbf{0.30}} &
  0.24 \\ \hline
trial\_T20190906\_234735\_610018-1 &
  \multicolumn{1}{c|}{\textbf{0.24}} &
  None &
  \multicolumn{1}{c|}{\textbf{0.00}} &
  None &
  \multicolumn{1}{c|}{\textbf{0.46}} &
  None &
  \multicolumn{1}{c|}{\textbf{0.46}} &
  None \\ \hline
trial\_T20190907\_200154\_378982-1 &
  \multicolumn{1}{c|}{\textbf{0.37}} &
  None &
  \multicolumn{1}{c|}{\textbf{0.27}} &
  None &
  \multicolumn{1}{c|}{\textbf{0.63}} &
  0.19 &
  \multicolumn{1}{c|}{0.20} &
  \textbf{0.44} \\ \hline
trial\_T20190907\_114323\_767231-2 &
  \multicolumn{1}{c|}{\textbf{0.00}} &
  None &
  \multicolumn{1}{c|}{0.10} &
  \textbf{0.42} &
  \multicolumn{1}{c|}{\textbf{0.47}} &
  0.26 &
  \multicolumn{1}{c|}{\textbf{0.30}} &
  0.09 \\ \hline
trial\_T20190906\_234735\_610018-2 &
  \multicolumn{1}{c|}{\textbf{0.41}} &
  None &
  \multicolumn{1}{c|}{0.07} &
  \textbf{0.23} &
  \multicolumn{1}{c|}{\textbf{0.41}} &
  None &
  \multicolumn{1}{c|}{\textbf{0.61}} &
  0.06 \\ \hline
trial\_T20190909\_082934\_483899-1 &
  \multicolumn{1}{c|}{0.31} &
  \textbf{0.47} &
  \multicolumn{1}{c|}{\textbf{0.31}} &
  None &
  \multicolumn{1}{c|}{\textbf{0.29}} &
  0.19 &
  \multicolumn{1}{c|}{\textbf{0.39}} &
  None \\ \hline
trial\_T20190909\_100946\_496614-1 &
  \multicolumn{1}{c|}{\textbf{0.27}} &
  None &
  \multicolumn{1}{c|}{\textbf{0.37}} &
  None &
  \multicolumn{1}{c|}{\textbf{0.37}} &
  0.21 &
  \multicolumn{1}{c|}{\textbf{0.47}} &
  0.17 \\ \hline
trial\_T20190907\_200154\_378982-2 &
  \multicolumn{1}{c|}{\textbf{0.55}} &
  None &
  \multicolumn{1}{c|}{\textbf{0.70}} &
  None &
  \multicolumn{1}{c|}{\textbf{0.65}} &
  0.20 &
  \multicolumn{1}{c|}{\textbf{0.65}} &
  0.18 \\ \hline
trial\_T20190909\_082934\_483899-2 &
  \multicolumn{1}{c|}{\textbf{0.14}} &
  None &
  \multicolumn{1}{c|}{\textbf{0.14}} &
  None &
  \multicolumn{1}{c|}{\textbf{0.42}} &
  0.14 &
  \multicolumn{1}{c|}{\textbf{0.29}} &
  0.14 \\ \hline
trial\_T20190909\_012550\_586494-1 &
  \multicolumn{1}{c|}{\textbf{0.30}} &
  None &
  \multicolumn{1}{c|}{\textbf{0.30}} &
  None &
  \multicolumn{1}{c|}{\textbf{0.29}} &
  0.05 &
  \multicolumn{1}{c|}{\textbf{0.43}} &
  0.08 \\ \hline
trial\_T20190909\_082934\_483899-3 &
  \multicolumn{1}{c|}{\textbf{0.29}} &
  None &
  \multicolumn{1}{c|}{\textbf{0.29}} &
  None &
  \multicolumn{1}{c|}{\textbf{0.57}} &
  0.21 &
  \multicolumn{1}{c|}{\textbf{0.57}} &
  0.10 \\ \hline
trial\_T20190909\_100946\_496614-2 &
  \multicolumn{1}{c|}{\textbf{0.55}} &
  None &
  \multicolumn{1}{c|}{\textbf{0.35}} &
  None &
  \multicolumn{1}{c|}{\textbf{0.65}} &
  0.34 &
  \multicolumn{1}{c|}{\textbf{0.65}} &
  0.42 \\ \hline
trial\_T20190906\_234735\_610018-3 &
  \multicolumn{1}{c|}{\textbf{0.54}} &
  None &
  \multicolumn{1}{c|}{\textbf{0.41}} &
  0.35 &
  \multicolumn{1}{c|}{\textbf{0.48}} &
  None &
  \multicolumn{1}{c|}{\textbf{0.61}} &
  None \\ \hline
trial\_T20190909\_012550\_586494-2 &
  \multicolumn{1}{c|}{\textbf{0.20}} &
  None &
  \multicolumn{1}{c|}{\textbf{0.20}} &
  None &
  \multicolumn{1}{c|}{\textbf{0.29}} &
  None &
  \multicolumn{1}{c|}{\textbf{0.34}} &
  None \\ \hline
trial\_T20190907\_114323\_767231-3 &
  \multicolumn{1}{c|}{\textbf{0.00}} &
  None &
  \multicolumn{1}{c|}{\textbf{0.10}} &
  None &
  \multicolumn{1}{c|}{\textbf{0.30}} &
  0.17 &
  \multicolumn{1}{c|}{\textbf{0.30}} &
  0.13 \\ \hline
trial\_T20190907\_114323\_767231-4 &
  \multicolumn{1}{c|}{\textbf{0.50}} &
  None &
  \multicolumn{1}{c|}{0.00} &
  \textbf{0.60} &
  \multicolumn{1}{c|}{\textbf{0.20}} &
  0.12 &
  \multicolumn{1}{c|}{\textbf{0.50}} &
  0.23 \\ \hline
trial\_T20190909\_004531\_429065-2 &
  \multicolumn{1}{c|}{\textbf{0.23}} &
  None &
  \multicolumn{1}{c|}{0.23} &
  \textbf{0.47} &
  \multicolumn{1}{c|}{\textbf{0.29}} &
  0.23 &
  \multicolumn{1}{c|}{\textbf{0.44}} &
  0.15 \\ \hline
trial\_T20190909\_193045\_208933-1 &
  \multicolumn{1}{c|}{\textbf{0.35}} &
  None &
  \multicolumn{1}{c|}{\textbf{0.35}} &
  None &
  \multicolumn{1}{c|}{\textbf{0.55}} &
  0.24 &
  \multicolumn{1}{c|}{\textbf{0.40}} &
  0.19 \\ \hline
trial\_T20190909\_193045\_208933-2 &
  \multicolumn{1}{c|}{\textbf{0.35}} &
  None &
  \multicolumn{1}{c|}{\textbf{0.45}} &
  0.33 &
  \multicolumn{1}{c|}{\textbf{0.65}} &
  0.27 &
  \multicolumn{1}{c|}{\textbf{0.30}} &
  0.20 \\ \hline
trial\_T20190907\_114323\_767231-5 &
  \multicolumn{1}{c|}{\textbf{0.23}} &
  None &
  \multicolumn{1}{c|}{\textbf{0.23}} &
  None &
  \multicolumn{1}{c|}{\textbf{0.33}} &
  0.23 &
  \multicolumn{1}{c|}{\textbf{0.33}} &
  None \\ \hline
trial\_T20190910\_173916\_331859-2 &
  \multicolumn{1}{c|}{\textbf{0.14}} &
  None &
  \multicolumn{1}{c|}{\textbf{0.07}} &
  None &
  \multicolumn{1}{c|}{\textbf{0.36}} &
  0.11 &
  \multicolumn{1}{c|}{\textbf{0.36}} &
  0.12 \\ \hline
trial\_T20190909\_004531\_429065-3 &
  \multicolumn{1}{c|}{\textbf{0.37}} &
  None &
  \multicolumn{1}{c|}{\textbf{0.23}} &
  None &
  \multicolumn{1}{c|}{\textbf{0.37}} &
  None &
  \multicolumn{1}{c|}{\textbf{0.43}} &
  0.21 \\ \hline
trial\_T20190907\_200154\_378982-3 &
  \multicolumn{1}{c|}{\textbf{0.20}} &
  None &
  \multicolumn{1}{c|}{\textbf{0.00}} &
  None &
  \multicolumn{1}{c|}{\textbf{0.00}} &
  None &
  \multicolumn{1}{c|}{\textbf{0.55}} &
  0.04 \\ \hline
trial\_T20190910\_173916\_331859-3 &
  \multicolumn{1}{c|}{\textbf{0.24}} &
  None &
  \multicolumn{1}{c|}{0.31} &
  \textbf{0.39} &
  \multicolumn{1}{c|}{\textbf{0.43}} &
  0.16 &
  \multicolumn{1}{c|}{\textbf{0.43}} &
  0.20 \\ \hline
trial\_T20190907\_114323\_767231-6 &
  \multicolumn{1}{c|}{\textbf{0.00}} &
  None &
  \multicolumn{1}{c|}{0.00} &
  \textbf{0.39} &
  \multicolumn{1}{c|}{\textbf{0.20}} &
  None &
  \multicolumn{1}{c|}{\textbf{0.40}} &
  0.30 \\ \hline
trial\_T20190909\_193045\_208933-3 &
  \multicolumn{1}{c|}{0.00} &
  \textbf{0.26} &
  \multicolumn{1}{c|}{0.00} &
  \textbf{0.35} &
  \multicolumn{1}{c|}{\textbf{0.20}} &
  None &
  \multicolumn{1}{c|}{\textbf{0.30}} &
  0.21 \\ \hline
trial\_T20190909\_012550\_586494-3 &
  \multicolumn{1}{c|}{\textbf{0.31}} &
  None &
  \multicolumn{1}{c|}{\textbf{0.24}} &
  None &
  \multicolumn{1}{c|}{\textbf{0.21}} &
  0.18 &
  \multicolumn{1}{c|}{\textbf{0.21}} &
  0.19 \\ \hline
\end{tabular}
\label{tab:results}
\end{table*}
An example of ground truth plan taken from the ALFRED dataset for the task ``cool the tomato in the refrigerator, put it by the toaster'' is the following:
\\ \\
\begin{minipage}{\columnwidth}
\begin{small}
\begin{itemize}[noitemsep]
    \item \texttt{Turn right, walk across the room to the counter on the right side of the stove.}
    \item \texttt{Grab the tomato on the counter.}
    \item \texttt{Turn around, walk to the refrigerator.}
    \item \texttt{Put the tomato in, close the refrigerator, wait a while, open the refrigerator, take the tomato out, close the refrigerator.}
    \item \texttt{Turn around, walk to the stove.}
    \item \texttt{Put the tomato down on the counter in front of the toaster.}
\end{itemize}
\end{small}
\vspace{0.5mm}
\end{minipage}

\noindent The corresponding predicted plan by our architectur for the same task is:

\noindent\begin{minipage}{\columnwidth}
\vspace{1mm}
\begin{small}
\begin{itemize}[noitemsep]
    \item \texttt{Walk to the kitchen island, which is in front of the kitchen counter.}
    \item \texttt{Pick up the tomato, which is on the kitchen island.}
    \item \texttt{Walk to the refrigerator, which is to the left of the kitchen island.}
    \item \texttt{Open the refrigerator door.}
    \item \texttt{Place the tomato inside the refrigerator.}
    \item \texttt{Close the refrigerator door.}
    \item \texttt{Wait for the tomato to cool.}
    \item \texttt{Open the refrigerator door.}
\end{itemize}
\end{small}
\end{minipage}
\noindent\begin{minipage}{\columnwidth}
\begin{small}
\begin{itemize}[noitemsep]
    \item \texttt{Take the cooled tomato out of the refrigerator.}
    \item \texttt{Close the refrigerator door.}
    \item \texttt{Walk to the kitchen counter, which is to the right of the refrigerator.}
    \item \texttt{Place the cooled tomato by the toaster, which is on the kitchen counter.}
\end{itemize}
\end{small}
\vspace{0.5mm}
\end{minipage}

The generated plan and the ground truth plan, compared with both metrics, obtain a 50\% percentage of similarity with PG2S compared to a 30\% percentage by KAS. The most notable distinction between the two plans is the navigation sequence required to reach the object in question. In the ground truth plan, the kitchen objects are reached differently than the predicted plan. Additionally, there is a notable difference in the number of actions required to cool a tomato in the refrigerator. In the predicted plan, there are seven steps, while in the ground truth, there is only one. Then, the predicted plan was corrupted by replacing \textit{kitchen} $\rightarrow$ \textit{bathroom} and \textit{tomato} $\rightarrow$ \textit{bottle}, and we verified the drop of similarity from 50\% to 25\% in PG2S, while KAS decreased from 30\% to 26\%. In this case, the PG2S similarity is halved, while in KAS the modification of the plan does not affect much. The step-by-step comparison performed by KAS does not allow an analysis of whether the goals are carried out during the planning; moreover, the cross-comparison of words loses sight of the semantic content. Table \ref{tab:results_corrupted} shows how the similarity score drops after corrupting the plan by changing the objects due to the PG2S semantic search. The plan available presents an id "trail\_ID"  associated with a ground truth plan and a goal that can be found in the G-Planet dataset.\footnote{\url{https://huggingface.co/datasets/yuchenlin/G-PlanET/viewer/default}}  The metric does not evaluate the success score of the plan but compares the steps with a semantic evaluation. Table \ref{tab:results_corrupted} also illustrates the scores obtained by comparing the plans obtained with KAS and PG2S before corruption. In each case, the degree of similarity obtained is superior to that of KAS.
Both metrics are used in the following section to evaluate the plan correction.
\begin{figure}[t!]
    \centering
    \includegraphics[width=1.0\columnwidth]{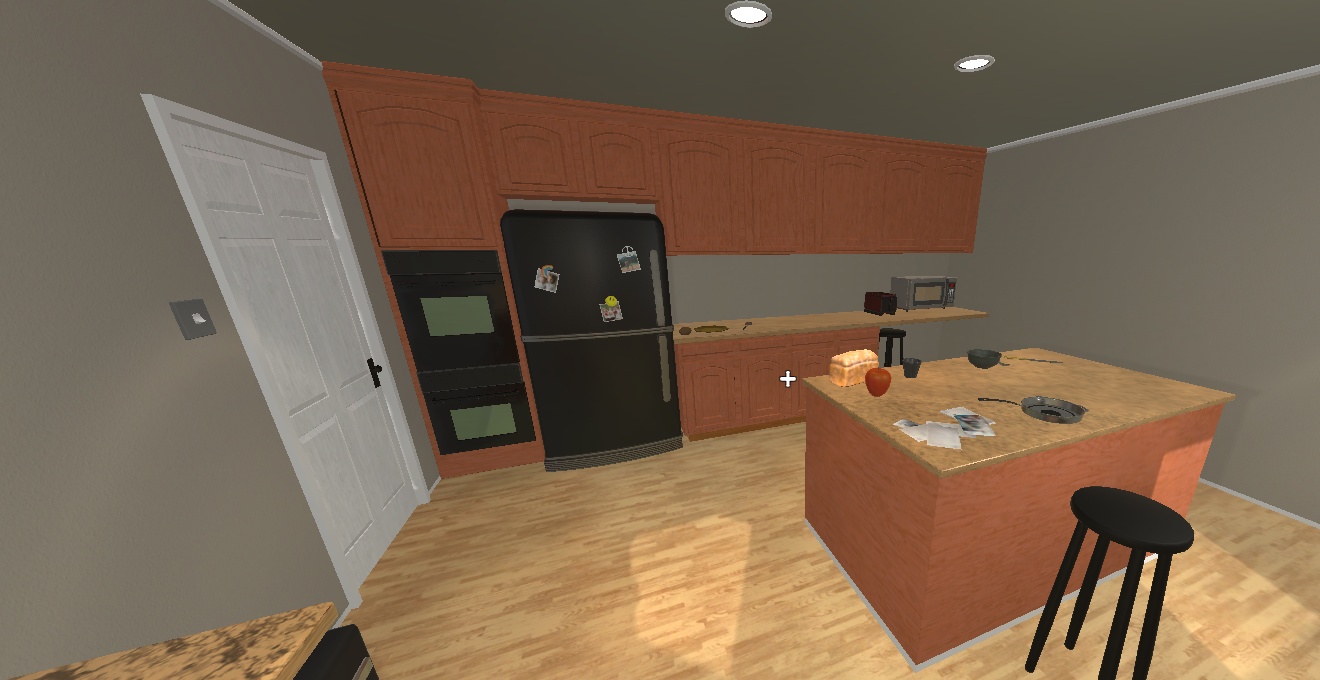}
    \caption{One of the scenes used for the experimental tests. A screenshot from AI2Thor is used to perform the planning.}
    \label{fig:example}
\end{figure}
\subsection{Evaluation of our Architecture}
To evaluate the presented methodology, we have chosen 
ten different rooms of an apartment, such as a living room, a kitchen, and a toilet. Frames were captured for each room as in the example in Fig. \ref{fig:example} which depicts a kitchen. The complexity of generating a plan is evident, given that an entire scene is represented by a single image and that some of the objects needed can be quite small. Our tests demonstrated that even in complex situations, the VLM is capable of identifying objects and perceiving their relationships, allowing it to define a correct plan.
The ten environments chosen allow us to obtain thirty tasks to perform and, for each of these plans, we have obtained the plan using four approaches: two using a single-agent architecture and two using a multi-agent architecture. In both single-agent and multi-agent evaluations, the plan was obtained using a table describing the environment rather than a single image.

Table \ref{tab:results} presents the results, highlighting instances where the KAS metric fails, resulting in \textit{None} values. This failure occurs because the KAS metric cannot evaluate plans of different lengths, which was common in the ``with table'' setups.

The results show how using a single image the architecture generates a plan similar to the ground truth plan. Furthermore, we demonstrated to obtain improved results in multi-agent architecture using a single image.


\section{Discussion}
The current state of the art involves the use of traditional Success Rate metrics to evaluate a plan, where the plan is considered correct in cases where execution leads to the desired outcome. 
However, this metric is not sufficient or suitable for all cases where the correctness of a task execution plan is to be analyzed. In particular, in cases where the plan is complicated, it should be evaluated before execution to avoid damage to the environment or simply unsuccessful executions and ensure that time and resources are not wasted in a new execution.
The advent of LLMs has made it possible to easily generate plans that previously required model training or other more complex techniques. Given that these models can' hallucinate' or generate incorrect responses, there could be errors present. Therefore, these inaccuracies could lead to failures when evaluating them based on success rates.
Our work seeks to define a new PG2S metric for plan evaluation based only on natural language processing while avoiding execution of the obtained plan to ensure the correctness of results. 
Although the presented metric can only provide a limited evaluation in cases where the final goal is not detailed enough, to the best of our knowledge, PG2S is the first contribution that addresses the problem in a way as general as possible. This paves the way for novel approaches where language processing techniques are adopted for the plan evaluation task. Future advances may improve the presented metric.

\section{Conclusion}
In this paper, we, first, introduced a multi-agent planning framework that leverages the capabilities of Visual Language Models (VLMs) to improve planning for embodied agents without the need for pre-encoded environmental data structures. Our approach simplifies the input requirements by utilizing a single environmental image and also enhances the adaptability and effectiveness of the planning process through a multi-agent system. This innovation addresses the limitations of traditional models that rely heavily on structured data, providing a more flexible and dynamic planning mechanism that is particularly effective in unstructured, real-world scenarios.

The empirical results, validated using the ALFRED dataset, demonstrate the efficacy of our approach, especially when compared to existing metrics like the KAS metric. We, then, introduce a new metric for the plan evaluation. The newly proposed PG2S metric, which assesses planning quality based on semantic understanding rather than strict action order, has shown superior performance in capturing the variations of plan execution.

The presented approach can address some of the current limitations in embodied agent planning and can open future research in the application of VLMs and multi-agent systems. Future studies might explore the scalability of our approach to more complex multi-agent environments and the integration of more diverse modalities to enhance the agents' understanding of their operational contexts. PG2S explores novel possibilities in the plan evaluation, focusing on semantic integrity rather than strict action sequencing. We believe that the research community can take advantage of the proposed approach, considering semantic coherence as a critical component of plan success, especially in applications requiring high reliability and safety.


\begin{ack}
This work has been carried out while Francesco Argenziano and Michele Brienza were enrolled in the Italian National Doctorate on Artificial Intelligence run by Sapienza University of Rome. \\
This work has been partially supported by PNRR MUR project PE0000013-FAIR. The research reported in the paper was partially supported by the project “Tech4You (ECS00000009) - Spoke 6”, under the NRRP MUR program funded by the NextGenerationEU.

\end{ack}



\bibliography{m2337}

\end{document}